\documentclass[letterpaper, 10 pt, conference, table]{ieeeconf}  %

\IEEEoverridecommandlockouts                              %
\makeatletter
\let\NAT@parse\undefined
\makeatother

\PassOptionsToPackage{usenames, dvipsnames}{xcolor}
\usepackage[usenames,dvipsnames,rgb]{xcolor}

\usepackage{microtype}

\newcommand*\linkcolours{ForestGreen}
\usepackage{subcaption}
\usepackage{times}
\usepackage{graphicx}
\usepackage{amssymb}
\usepackage{gensymb}
\usepackage{amsmath}
\usepackage{breakurl}

\usepackage{url,hyperref}
\usepackage{booktabs}
\hypersetup{
colorlinks,
linkcolor=\linkcolours,
citecolor=\linkcolours,
filecolor=\linkcolours,
urlcolor=\linkcolours}

\usepackage{algpseudocode}
\usepackage{algorithm}

\algnewcommand\algorithmicforeach{\textbf{for each}}
\algdef{S}[FOR]{ForEach}[1]{\algorithmicforeach\ #1\ \algorithmicdo}

\usepackage[labelfont={bf},font=small]{caption}
\usepackage[none]{hyphenat}

\usepackage{mathtools, cuted}

\usepackage[noadjust, nobreak]{cite}

\usepackage{tabularx}
\usepackage{amsmath}

\usepackage{float}

\usepackage{pifont}%

\newcolumntype{Y}{>{\centering\arraybackslash}X}

\usepackage[]{placeins}

\usepackage{placeins}

\usepackage{tikz}

\usepackage[framemethod=tikz]{mdframed}

\usepackage{afterpage}

\usepackage{stfloats}

\usepackage{atbegshi}
\newcommand{\handlethispage}{}
\newcommand{\discardpagesfromhere}{\let\handlethispage\AtBeginShipoutDiscard}
\newcommand{\keeppagesfromhere}{\let\handlethispage\relax}
\AtBeginShipout{\handlethispage}

\usepackage{comment}
\usepackage{bm}

\def\Nc{\mathcal{N}}
\def\Ic{\mathcal{I}}
\def\Ac{\mathcal{A}}

\def\Dc{\mathcal{D}}
\def\Ec{\mathcal{E}}
\def\Mc{\mathcal{M}}

\usepackage[T1]{fontenc}
\usepackage[latin9]{inputenc}
\usepackage{babel}
\usepackage{collcell}
\usepackage{hhline}
\usepackage{pgf}
\usepackage{multirow}

\def\colorModel{hsb} %
\newcommand\ColCell[1]{
	\pgfmathparse{#1<50?1:0}  %
	\ifnum\pgfmathresult=0\relax\color{white}\fi
	\pgfmathsetmacro\compA{0}      %
	\pgfmathsetmacro\compB{#1/100} %
	\pgfmathsetmacro\compC{1}      %
	\edef\x{\noexpand\centering\noexpand\cellcolor[\colorModel]{\compA,\compB,\compC}}\x #1
}
\newcolumntype{E}{>{\collectcell\ColCell}m{0.4cm}<{\endcollectcell}}  %
\newcommand*\rot{\rotatebox{90}}

\DeclareMathOperator*{\argmax}{argmax}

\title{\LARGE \bf
A Reinforcement Learning Approach for Rebalancing Electric Vehicle Sharing Systems
}

\author{
Aigerim Bogyrbayeva$^\dag$, Sungwook Jang$^\flat$, Ankit Shah$^\dag$, Young Jae Jang$^\flat$, Changhyun Kwon$^\dag$%
\thanks{$^\dag$Department of Industrial and Management Systems Engineering, University of South Florida, Tampa, Florida, USA, Email: {\tt\footnotesize\{aigerimb, ankitshah, chkwon\}@usf.edu}}%
\thanks{$^\flat$Department of Industrial and Systems Engineering, KAIST, Daejeon, South Korea, Email: {\tt\footnotesize\{jedi829, yjang\}@kaist.ac.kr}}%
}

\begin{document}

\maketitle
\thispagestyle{plain}
\pagestyle{plain}

\begin{abstract}
This paper proposes a reinforcement learning approach for nightly offline rebalancing operations in free-floating electric vehicle sharing systems (FFEVSS). Due to sparse demand in a network, FFEVSS requires relocation of electrical vehicles (EVs) to charging stations and demander nodes, which is typically done by a group of drivers. A shuttle is used to pick up and drop off drivers throughout the network. The objective of this study is to solve the shuttle routing problem to finish the rebalancing work in minimal time. We consider a reinforcement learning framework for the problem, in which a central controller determines the routing policies of a fleet of multiple shuttles. We deploy a policy gradient method for training recurrent neural networks and compare the obtained policy results with heuristic solutions. Our numerical studies show that unlike the existing solutions in the literature, the proposed methods allow solving the general version of the problem with no restrictions on the urban EV network structure and charging requirements of EVs. Moreover, the learned policies offer a wide range of flexibility, resulting in a significant reduction in the time needed to rebalance the network.
\end{abstract}


\section{Introduction}

The advent of electric vehicles (EVs) and car-sharing services provides a sustainable option to move people and goods across dense urban areas. Car sharing services with EVs have the potential to increase the utilization of resources and offer a unique opportunity to the urban population in the form of free-floating EV sharing systems (FFEVSS). With the FFEVSS, examples of which include companies such as car2go \cite{car2go} and WeShare \cite{vw}, customers no longer need to own a vehicle and can conveniently pick up/drop off any EV, on-demand, from the parking lots of designated service areas. However, there are some critical operational challenges to bring this on-demand service into the mainstream.

Before the start of the day, an operating company needs to relocate EVs to the ideal demand locations to establish a supply-demand balance in the system. 
Furthermore, to provide a certain level of service, EVs need to be charged before they can be used by the customers. There are two major issues: (i) there exists a sparse demand in the service area network, and hence it is not trivial to find the ideal locations to relocate the EVs; and (ii) there needs to be an efficient routing plan to drop off the drivers for picking up the EVs and taking the EVs to the charging stations for charging, and then pick up the drivers from their respective locations \cite{car2go_news}. %
It is evident that without efficient solutions for the above complex and costly operational challenges \cite{chianese2017one}, the sustainable existence of the FFEVSS is uncertain.

\begin{figure}
	\centering 
	\includegraphics[scale=0.3]{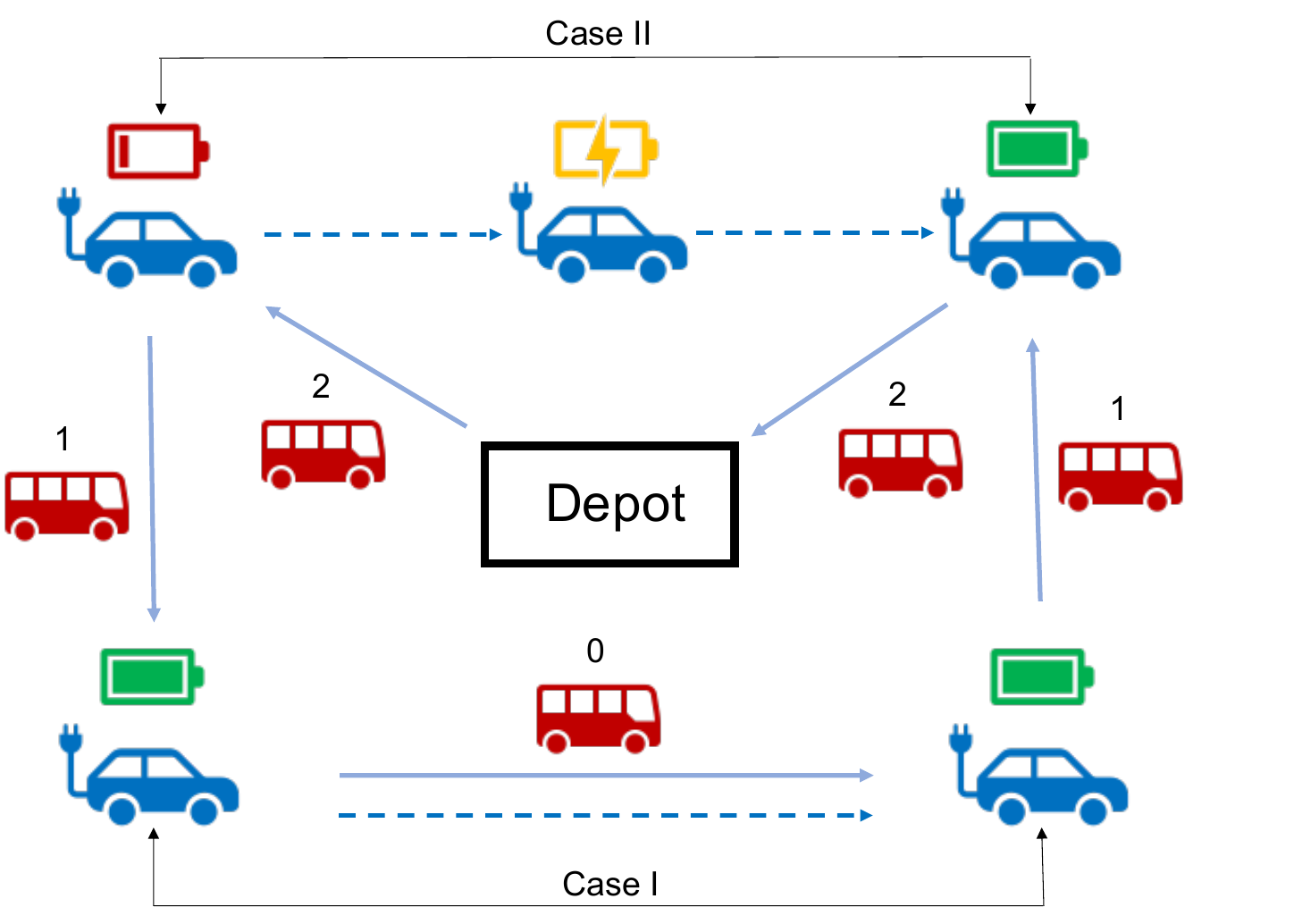}
	\caption{The overview of the rebalancing problem of FFEVSS, with a single shuttle and 2 drivers. The numbers indicate the number of drivers in the shuttle. Solid and dashed lines represent the routes of the shuttle and EVs, respectively. Cases I and II refer to relocation of EVs without and with charging, respectively.}
	\label{fig:relocate_overview}
\end{figure}

We consider a static, nightly rebalancing problem similar to \cite{kypriadis2018minimum, santos2017vehicle, folkestad2020optimal, zulqar}, where a group of drivers is used to relocate and recharge the EVs based on the predicted demand for the next day, assuming the utilization level of FFEVSS is minimal.
As shown in Figure \ref{fig:relocate_overview}, shuttles are used to support the movements of drivers.
In this setting, rebalancing operations require two key decisions to be made: (i) how to route shuttles to pick up and drop off the drivers (shuttle routing decision) and (ii) where to charge and relocate each of the EVs (EV relocation decision). 
In this paper, focusing on solving the shuttle routing decision problem, we propose a reinforcement learning (RL) approach, in which the EV relocation decisions are made by a rule-based approach.

The proposed RL approach possesses several advantages compared to optimization-based approaches.
First, unlike solutions coming from the static optimization techniques such as \cite{folkestad2020optimal, zulqar}, which need to be re-solved each time an input changes, the RL agent learns robust solutions that can be applied to any input coming from the same distribution \cite{bello2016neural}.
Second, while static optimization approaches can take significant time to solve a problem, a trained RL agent can be invoked to produce quality solutions instantaneously.
Third, many practical considerations can be flexibly incorporated within the simulator in the training phase.

The shuttle routing to rebalance FFEVSS with its variety of trade-offs is not a trivial problem.
For instance, as depicted in Figure \ref{fig:charge_resuse}, one may allow or disallow the reuse of charging stations in the derivation of solutions.
The former choice offers more flexibility, but it also increases the complexity of exploring solutions.
Therefore, the existing methods do not allow the reuse of the charging stations \cite{zulqar}.
On the other hand, such a choice results in opportunity loss.

Another trade-off is depicted in Figure \ref{fig:trade_off}, where the first supplier node has an EV that needs to be recharged while the second supplier has an EV with a sufficient charging level.
Then one needs to balance between traveling time and waiting time when routing a shuttle to supplier nodes.
The complexity of such routing decisions increases with the network size, the network structure, and the number of shuttles and drivers deployed.
Hence, it may not be possible to explore potential solutions with human-driven heuristics efficiently. %
With the proven ability of neural networks in recognizing patterns in graph-based representations, the utilization of neural network architecture with the proposed RL approach will provide better approximations and assist in obtaining efficient solutions that can be generalized.

\tikzset{
	intercepted_circle/.style={circle, draw=black, fill=black, inner sep=0pt, minimum size=8pt},
	unintercepted_circle/.style={circle, draw=black, fill=white, inner sep=0pt, minimum size=12pt}
}

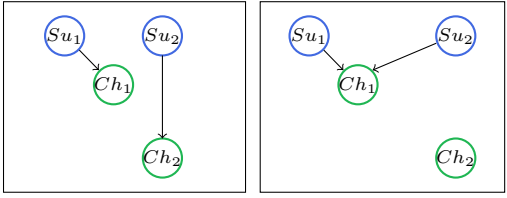
\begin{figure}
	\centering
	\begin{tikzpicture}[scale=0.65]
	\definecolor{green}{RGB}{31,182,83}
	\definecolor{red}{RGB}{203, 65, 84}
	\definecolor{blue}{RGB}{65, 105, 225}
	\draw[blue, thick, text=black] (0, 0) circle [radius=0.4] node {\scriptsize{$Su_{1}$}};
	\draw[green, thick, text=black] (1, -1) circle [radius=0.4] node {\scriptsize{$Ch_{1}$}};
	\draw [->] (0.3, -0.29)-- (0.7, -0.7) ;
	\draw[blue, thick, text=black] (2, 0) circle [radius=0.4] node {\scriptsize{$Su_{2}$}};
	\draw[green,  thick, text=black] (2, -2.5) circle [radius=0.4] node {\scriptsize{$Ch_{2}$}};
	\draw [->] (2, -0.4)-- (2, -2.1) ;
	\draw[black] (-1.25, 0.7)  rectangle (3.7, -3.2) ;
	\draw[thick, blue, text=black] (5, 0) circle [radius=0.4] node {\scriptsize{$Su_{1}$}};
	\draw[thick, green, text=black] (6, -1) circle [radius=0.4] node {\scriptsize{$Ch_{1}$}};
	\draw [->] (5.3, -0.29)-- (5.7, -0.7) ;
	\draw[thick, blue, text=black] (8, 0) circle [radius=0.4] node {\scriptsize{$Su_{2}$}};
	\draw[thick, green, text=black] (8, -2.5) circle [radius=0.4] node {\scriptsize{$Ch_{2}$}};
	\draw [->] (7.6, -0.15)-- (6.3, -0.7) ;
	\draw[black] (4, 0.7)  rectangle (9, -3.2) ;
	\end{tikzpicture}
	\caption{Assign supplier-charger pairs or reuse charger nodes? $Ch$ and $Su$ denotes charger and supplier nodes respectively.}
	\label{fig:charge_resuse}
\end{figure}

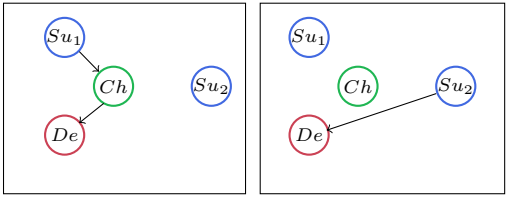
\begin{figure}
	\centering
	\begin{tikzpicture}[scale=0.65]
	\definecolor{green}{RGB}{31,182,83}
	\definecolor{red}{RGB}{203, 65, 84}
	\definecolor{blue}{RGB}{65, 105, 225}
	\draw[thick, blue, text=black] (0, 0) circle [radius=0.4] node {\scriptsize{$Su_{1}$}};
	\draw[thick, green, text=black] (1, -1) circle [radius=0.4] node {\scriptsize{$Ch$}};
	\draw [->] (0.3, -0.29)-- (0.7, -0.7) ;
	\draw[thick, red, text=black] (0, -2) circle [radius=0.4] node {\scriptsize{$De$}};
	\draw[thick, blue, text=black] (3, -1) circle [radius=0.4] node {\scriptsize{$Su_{2}$}};
	\draw [->] (0.8, -1.35)-- (0.3, -1.75) ;
	\draw[black] (-1.25, 0.7)  rectangle (3.7, -3.2) ;
	\draw[thick, blue, text=black] (5, 0) circle [radius=0.4] node {\scriptsize{$Su_{1}$}};
	\draw[thick, green, text=black] (6, -1) circle [radius=0.4] node {\scriptsize{$Ch$}};
	\draw[thick, blue, text=black] (8, -1) circle [radius=0.4] node {\scriptsize{$Su_{2}$}};
	\draw[thick, red, text=black] (5, -2) circle [radius=0.4] node {\scriptsize{$De$}};
	\draw [->] (7.6, -1.15)-- (5.36, -1.9) ;
	\draw[black] (4, 0.7)  rectangle (9, -3.2) ;
	\end{tikzpicture}
	\caption{How to balance traveling time and waiting time trade-off? $De$, $Ch$ and $Su$ denotes demander, charger and supplier nodes respectively.}
	\label{fig:trade_off}
\end{figure}

\begin{figure}\centering
	\begin{tikzpicture}[scale=0.8]
	\definecolor{green}{RGB}{31,182,83}
	\definecolor{red}{RGB}{203, 65, 84}
	\definecolor{blue}{RGB}{65, 105, 225}
	\draw[black, text=black] (0, 0) circle [radius=0.3] node {$S_0$};
	\draw (3, 0) circle [radius=0.3] node {$S_1$};
	\draw (6, 0) circle [radius=0.3] node {$S_2$};
	\draw [->] (0.3, 0)-- node[above,font=\scriptsize] {$a_0: \mathrm{SR}_0$} (2.7, 0) ;
	\draw [->] (3.3, 0) -- node[above,font=\scriptsize] {$a_1: \mathrm{SR}_1$} node[below,font=\scriptsize] {$\mathrm{EVR}_0$}(5.7, 0);
	\draw [->] (6.3, 0) -- node[above,font=\scriptsize] {$a_2: \mathrm{SR}_2$} node[below,font=\scriptsize] {$\mathrm{EVR}_1$} (8.7, 0);
	\end{tikzpicture}
	\caption{State transitions: $\mathrm{SR}_i$ - shuttle routing decisions, $\mathrm{EVR}_i$ - EV relocation decisions, $a_i$ - selected action  }
	\label{fig:state_transitions}
\end{figure}
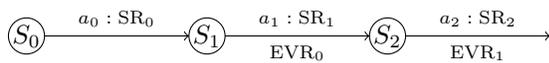

In recent years, there has been a surge of studies that apply reinforcement learning to solve various traditional vehicle routing problems (VRPs) \cite{nazari2018reinforcement,kalakanti2019rl,james2019online} with capacity constraints, time windows, or stochastic demand.
The shuttle routing problem, taken under this study, possesses significant differences with traditional VRPs.
First, in a VRP setting, nodes to visit (demand) are typically independent of the routing decisions. However, in the shuttle routing problem, the locations of drivers to be picked up are determined by preceding routing decisions. This highlights a \emph{strong interdependence} between demand and routing. 
Second, unlike VRP, the shuttle routing problem is characterized by \emph{delayed rewards}.
As shown in Figure \ref{fig:state_transitions}, the actual relocations of EVs from a node happen after the execution of shuttle routing to the node. As a result, we observe delayed rewards with respect to the shuttle routing decision only after EVs reach their designated nodes.
Such differences require a new approach to finding solutions for the shuttle routing problem.

We consider two settings of rebalancing FFEVSS. In the first setting, we focus on a single shuttle problem, where we train a single agent to learn routing policies. 
In the second setting, we aim to train a fleet of shuttles through single-agent reinforcement learning, where a central controller is responsible for routing multiple shuttles. In both cases, we deploy policy gradient methods along with recurrent neural networks for training.
The shuttle routing problem under both of the above-mentioned settings possesses significant challenges that prohibit the direct use of the existing solution methods. For instance, in routing a single shuttle, we must train an agent not only to find efficient routes, but at the same time maintain the feasibility of the solutions related to the precedence of the visiting nodes. As for routing the fleet of shuttles, we must promote learning policies to route multiple shuttles that will contribute to a common goal.

The main contributions of this study are as follows.
First, to the best of our knowledge, this study is the first to present an RL-based approach for handling \emph{multiple} vehicles \emph{explicitly} in the context of VRPs, while focusing on the shuttle routing problem for rebalancing the FFEVSS.
Second, within the RL framework, we propose the utilization of deep neural network architecture to process the complex and high dimensional observations from an urban service area network to help train the RL agent in its decision-making.
In particular, we adopt sequence-to-sequence models with an attention mechanism to fit the unique challenges of the rebalancing FFEVSS.
Third, we present a novel training algorithm to route efficiently a fleet of shuttles to rebalance FFEVSS by utilizing policy gradient methods. Our training algorithm does not require splitting an urban network into sub-clusters for each shuttle, but instead allows developing policies that efficiently utilize shuttles and drivers in a whole network.
Fourth, we develop a simulator to mimic real-world FFEVSS, which serves as the environment for training an RL-agent and allows efficient exploration of joint actions of multiple shuttles.

Unlike the solutions obtained using the methods from the literature, the empirical results obtained from this study show that the proposed method allows solving the general version of the problem with no restrictions on the urban network structure and charging levels of EVs.
The learned policies offer a wide range of flexibility, resulting in a significant reduction in the time needed to rebalance the network.

The remainder of the paper will proceed as follows. In Section \ref{sec:related} we provide an overview of relevant literature and outline the unique challenges of the rebalancing FFEVSS. In Section \ref{sec:statement} we present the problem formulation. In Section \ref{sec:rl} we introduce the proposed reinforcement learning model. In Section \ref{sec:na} we demonstrate the results of our computational studies. Lastly, in Section \ref{sec:conclusion} we provide concluding remarks.

\section{Related Work}\label{sec:related}

Even though the problem of rebalancing FFEVSS has been recognized as essential for their sustainable existence in the literature \cite{schulte2015decision, herrmann2014increasing}, most of the studies focus on high-level approaches to address the issue. One category of studies falls on incentive-based methods that aim to rebalance the system through influencing customer behavior \cite{weikl2013relocation}. Another set of papers study the deployment of personnel and offer rule-based high-level decision-making frameworks \cite{weikl2015practice, zhao2018integrated}. There are only a few studies that specifically focus on the shuttle routing problem to rebalance FFEVSS, thus offering detailed solutions for day-to-day operational challenges.

One of such studies is \cite{folkestad2020optimal}, which aims to solve both EV relocation and shuttle routing problems jointly. However, the proposed model does not enforce relocation of  EVs directly to demander nodes, but indeed permits leaving EVs in charger nodes. As a result, charger stations will be blocked and cannot be reused, requiring the postponing of charging for the remaining set of EVs.  Similarly, a recent study \cite{zulqar} presents novel approaches in addressing  EV relocation and shuttle routing problems simultaneously. Even though the study aims at relocating EVs directly to demander nodes, it assumes the abundance of charger stations in an urban network. Thus, again reusing charger stations is not considered, and the postponement of charging for EVs requiring it is allowed. Since charging infrastructure is often limited \cite{he2020charging}, the reuse of charging stations must be an integral part of solutions to rebalance FFEVSS in real-world urban networks.

Recently reinforcement learning approaches gained popularity to solve various problems in transportation, including fleet management and rebalancing in ride-hailing services \cite{shi2019operating, lin2018efficient, wen2017rebalancing, sadeghianpourhamami2018achieving}. 
However, none of the existing studies focus on FFEVSS specifically and do not address the unique issue of charging and relocation together.
For solving VRPs, deep reinforcement learning has been first applied in \cite{nazari2018reinforcement}, which utilizes sequence-to-sequence methods \cite{sutskever2014sequence} and an attention mechanism \cite{vinyals2015order}. Later \cite{kool2018attention} adopted the transformer model \cite{vaswani2017attention} to solve VRPs without recurrent neural networks. \cite{james2019online} proposes a novel model to solve online VRPs by utilizing neural combinatorial optimization and deep reinforcement learning. Similarly, \cite{zhao2020hybrid} presents a hybrid model that combines local search with an attention mechanism.  However, these studies focus on routing a single capacitated vehicle, where the main goal is to minimize the distance traveled.
While multiple loops of a single capacitated vehicle can be interpreted as multiple vehicles, this paper is the first to present explicit modeling of multiple vehicles within an RL framework.

Although this study also adopts sequence-to-sequence models with an attention mechanism similar to \cite{nazari2018reinforcement}, the significant differences in the nature of the rebalancing FFEVSS problem and VRP dictate the development of novel solution techniques.
For instance, in the given problem, shuttles need to leave a depot, drop off, pick up drivers who relocate EVs, and return to the depot, highlighting two sets of constraints. First, the precedence of visited nodes needs to be maintained when charging stations are visited after nodes with EVs and nodes that require EVs are visited after either charging stations or nodes with EVs. Second, the capacity constraint must be satisfied when nodes with EVs are visited only when there is a driver in a shuttle and nodes with drivers are visited only if there is seating available for a driver in the shuttle. In addition to feasibility constraints, since both charging and relocations of EVs are involved in the shuttle routing problem, only considering factors that affect the total distance traveled is not sufficient. Moreover, the dynamics of an urban network due to routing a shuttle is more complex compared to the VRP due to the delayed movements of EVs relocation.
Also, routing multiple shuttles requires a novel training algorithm. In particular, when several shuttles are present in an urban network and each of their movement influence the state of the network, we need a novel framework that enables the application of reinforcement learning tools based on Markov Decision Process (MDP).

\section{Problem Statement and Formulations}\label{sec:statement}

\subsection{Network}
Let us consider a network $\Nc$ consisting of $N$ number of nodes and a depot. We define a node as a supplier if it has an excess EV and a demander if it requires an EV. The network also has charger nodes. Each node in the network can store at most one EV. Depending on the charging levels of EVs there are two possibilities of the EVs relocation. In Case I, EVs are relocated from supplier nodes directly to demander nodes. In Case II, EVs first need to be taken to charger nodes, and after charging is complete, they need to be relocated to the demander nodes, as shown in Figure \ref{fig:relocate_overview}. We consider discrete charging levels of EVs, where a threshold-based rule is applied to decide whether to charge an EV or not. Also, a driver may wait at a charging station until an EV is fully charged or may head for the next activity.
We consider two settings of the problem when a single shuttle or a fleet of shuttles is deployed for rebalancing the system. We formulate the routing problem for a single shuttle as MDP and utilize a central controller to route a fleet of shuttles.

\tikzset{
	intercepted_circle/.style={circle, draw=black, fill=black, inner sep=0pt, minimum size=8pt},
	unintercepted_circle/.style={circle, draw=black, fill=white, inner sep=0pt, minimum size=12pt}
}

\subsection{ Multi-shuttle Routing as MDP}

Even though it is possible to formulate the routing of a fleet of shuttles using a multi-agent reinforcement learning framework, such an approach suffers from several drawbacks. Firstly, in the presence of several shuttles, each of which is treated as an autonomous agent, the stationary assumption of MDP is no longer valid \cite{bucsoniu2010multi}. In particular, in the presence of other agents in the environment, the Markovian property, which states that reward and current state only depends on individual action and previous state, does not hold. Therefore, a multi-agent reinforcement learning framework works under partially observable MDP \cite{lowe2017multi, gupta2017cooperative, foerster2018counterfactual}, when each agent can observe only a local view of the network \cite{oliehoek2016concise}. Then each agent can only visit nodes visible from its local view, which imposes significant restrictions on developing an efficient routing.
Secondly, it is challenging to train autonomous agents without making strong assumptions about constant communication between agents. For instance, if at the current time step one agent selects a node to visit, then such information must be shared among other agents to avoid the presence of several agents at the same node. Lastly, under a static network, when the state of the network is constant and well-known, a centralized approach will help navigate a fleet of shuttles efficiently.
Therefore, we formulate routing multiple shuttles to rebalance FFEVSS using a central controller responsible for making routing decisions of all shuttles. Then, we can formulate the problem using a single-agent reinforcement learning framework and MDP. We also note that the concept of multi-agent reinforcement learning and central controller is similar to decentralized control and centralized control in the transportation literature.

A fleet of shuttles with drivers leaves a depot and visits nodes in the network to relocate EVs from supplier nodes to demander nodes. Shuttles must return to a depot after fulfilling demand at all demander nodes and picking up all the drivers. These sequential decisions of a central controller for routing shuttles under uncertain demand (locations of drivers) can be formulated as a finite horizon MDP, where the future dynamics of the system depend only on the current state. We define the RL framework for the problem as tuple $\Mc ={\langle}X, \Ac, P, R, T {\rangle}$ representing states, actions, transition probabilities, reward function, and time horizon, respectively. The definitions are as follows:
\begin{itemize}
	\item $\Ic = \{1, ..., I\}$ is the set of $I$ shuttles that are controlled by a central controller;

	\item State set $X$ represents the network, where for each node it shows its location, the relative distance, the number of EVs, the number of drivers, the charging levels of EVs' and indicators for the expected transitions. We utilize binary vectors to indicate if there is an expected EV coming to a node. We denote state as $x_t$ at time $t$.

	\item $\Ac$ is the set of joint actions such that $\Ac_t = \Ac^1_t \times \Ac^2_t \times \cdots \times \Ac^{I}_t$, where $\Ac^i_t$ is the action set of shuttle $i$ at time $t$ and action $a^i_t$ indicates a node number to be visited next by shuttle $i$. Then a central controller's action set consists of joint actions of all shuttles, $\Ac_t$, at time $t$.

	\item Transition Probabilities function, $P$, determines state transitions probabilities $p(x_{t+1}|x_t, a_t)$ at time $t$  with respect to taken action $a_t$. In the given problem, transitions are deterministic but often delayed. After an action is taken, the relocations of EVs are scheduled. However, the actual state transitions related to the movements of EVs occur later, as shown in Figure \ref{fig:state_transitions}.

	\item All shuttles share a common reward $R$ and immediate reward $r_t$, which are assigned based on the joint actions of all shuttles at time $t$ denoted by $a_t$ and state $x_t$;

	\item Instead of defining the specific time value of $T$, we define one episode rollout for the problem based on the experiment outcomes. One episode is terminated either if all demander nodes are fulfilled and all drives are picked up back to a depot or if the total number of time steps exceeds the predefined maximum time steps, the value of which is set based on the size of a network.

	\item Each time step $t$ is determined by the earliest fulfilled action among all shuttles. Thus, each time step starts when a central controller takes an action and finishes whenever any action is fully executed.
\end{itemize}

The list of variables used can be found in Table \ref{tb:summary}.

\begin{table}[]
	\caption{A summary of variables.}
	\centering
	\label{tb:summary}
	\begin{tabular}{@{}llll@{}}
		\toprule
		$\Ic$ & the set of  shuttles  & 
		$\Ac$ & the set of joint actions \\
		$X$ & the state of network &
		$t_c$ & current clock time \\
		$\tau$ & traveling time &
		$w$ & waiting time \\
		$T$ & the max number of time steps&
		$r$ & immediate reward \\
		$R$ & total reward &
		$\theta_a$ & parameters of actor \\
		$\theta_c$ & parameters of critic &
		$\pi$ & routing policy \\
		$Y$ & the set of visited nodes &
		$n$ & node in network \\
		\bottomrule
	\end{tabular}
\end{table}

\section{Reinforcement Learning Model}\label{sec:rl}

\begin{figure*}[t]
	\centering 
	\includegraphics[width=4in]{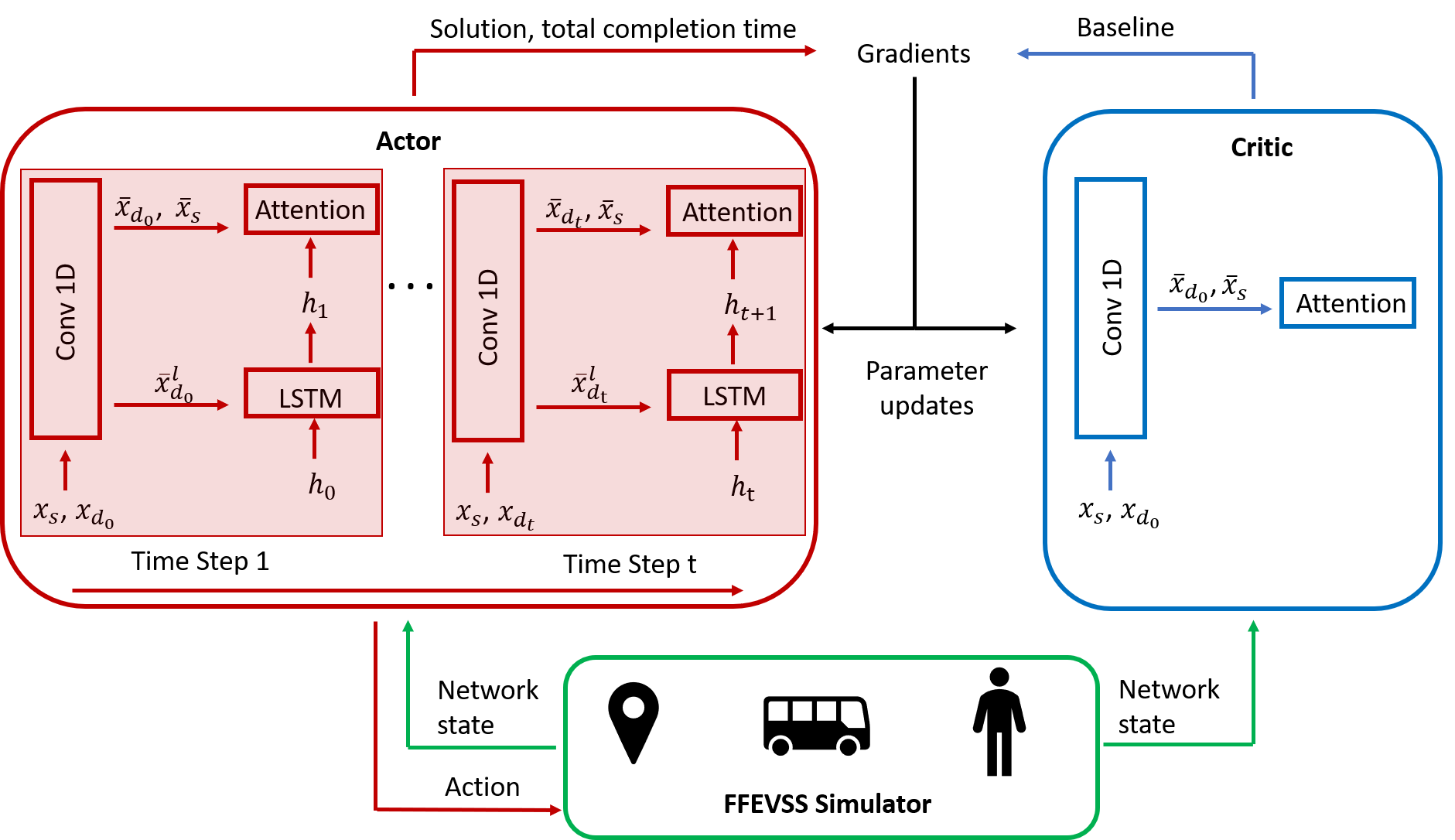}
	\caption{An overview of the reinforcement learning model.}
	\label{fig:rl_overview}
\end{figure*}

We adopt a policy gradient method, similar to those popularly used in routing problems \cite{nazari2018reinforcement, kool2018attention, james2019online}, to learn the complex routing policies of shuttles \emph{directly}.
In general, policy gradient methods consist of two separate networks: an actor and a critic.
The critic estimates a value function given a state according to which the actor's parameters are set to generate policies in the direction of improvement. %
We train an agent and a central controller to route a single shuttle and multiple shuttles in an urban network by simulating the FFEVSS environment. The simulator is developed to handle EV relocations through rule-based decisions and utilizing sequence-to-sequence models to generate policies. The overview of the model is shown in Figure \ref{fig:rl_overview}.

\subsection{The FFEVSS Simulator}
The main function of the FFEVSS simulator is to represent the dynamics in an urban network caused by movements of shuttles. There are immediate and delayed transitions related to routing shuttles. In an immediate update to the environment at each time step, we consider locations of shuttles, drivers, EVs, the number of drivers in a shuttle, and fulfillment of scheduled transitions either related to charging or relocation of EVs. Also, at each time step, we schedule transitions related to movements of EVs that have started but unfulfilled. In particular, starting at the current clock time $t_c=0$, we update the environment according to movements of a shuttle:
\[ t_c \gets
\begin{cases*}
t_c +\tau(n_{t-1}, n_{t}) & \text{if} $n_{t-1}\neq n_t$ \\
t_c + w_t & \text{if} $n_{t-1} = n_t$  \\
\end{cases*} \]
where $\tau$ represents traveling time between nodes visited by a shuttle at time $t-1$ and $t$ and $w_t$ denotes waiting time at node $n$. We define waiting time at node $n$ as the difference between the time when a delayed transition at node $n$ occurs and the time when a shuttle reaches node $n$. To account for delayed transitions, we introduce a time vector, which keeps track of remaining times until either EVs arrive at designated nodes or their charging completes. In the case of multiple shuttles, the environment is updated with the earliest movements of shuttles.

Another function of the FFEVSS simulator is to update a masking scheme according to the current state of the urban network. The masking scheme helps to maintain the feasibility of solutions related to the precedence of visited nodes and the number of drivers in a shuttle. Also, having an efficient masking scheme expedites the exploration of action space. We deploy the following masking scheme, where $\Ac_{t}= \emptyset$ stores the set of available nodes/actions to visit at time $t$ and the rest of the nodes are masked. For each $n \in \Nc$, we update:
\[ \Ac_{t} \gets
\begin{cases*}
 \Ac_{t} \cup \{n\} & \text{if}  $l_{t} > 0 \quad \text{and }n \in \Dc_{t} \cup \Ec_{t}$ \\
 \Ac_t \cup \{n\} & \text{if}  $l_{t} = 0 \quad \text{and } n \in \Dc_{t}$
\end{cases*} \]%
Here set $\Ec_t$ denotes nodes with an EV or nodes with the expected EV due to delayed transitions, set $\Dc_t$ denotes nodes with a driver or nodes with the expected drivers, and $l_t$ denotes the number of drivers in a shuttle at time $t$.

\subsection{EV relocation decisions}
As described earlier, our focus in this study is to solve the shuttle routing problem. Hence, we are using a rule-based approach for EVs' relocation decisions. The rule-based approach is as follows: every time a supplier node with an EV has a driver, that EV is relocated to the nearest available either charger or demander node. The decision of whether to relocate an EV to a demander or charger node is predetermined in the settings of a simulator. We apply a threshold-based rule; that is, if the charging level of an EV exceeds the threshold, then it can be directly relocated to a demander node or must be charged, otherwise.

We maintain a binary vector in the simulator to indicate if a charger node is available or not. This representation helps in deciding the relocation of an EV from a supplier node to an available charger node. We determine the closest available charger node by multiplying the binary vector by a time matrix that indicates time to travel among any pair of nodes.
To decide EVs' relocations from either supplier or charger nodes to demander nodes, we maintain a demand matrix that keeps track of demander nodes that still need an EV at time $t$. In particular, in the simulator, we store the time needed to move from all nodes to each demander node and increase those values to large numbers if a demander node is satisfied. Then, if an EV needs to be relocated to a demander node, we compute the minimum time from a node to the closest demander nodes.

\subsection{A sequence-to-sequence model for the shuttle routing problem}

Motivated by \cite{nazari2018reinforcement}, we propose using a sequence-to-sequence model for rebalancing FFEVSS, which typically consists of an encoder and a decoder. Given urban network $\Nc$, we aim to generate a sequence of nodes to be visited by either a shuttle or a fleet of shuttles. In other words, we are interested in learning the following conditional probability or parametrized policy $\pi_{\theta}$:

\begin{equation} \label{cond_prob}
\pi_{\theta}(Y_{T}| x_0) = \prod_{t=0}^{T-1} \phi(y_{t+1}| x_t, Y_t)
\end{equation}
In \eqref{cond_prob}, we let $x_t = \{x^1_t, \dots, x^N_t\}$, where $x^n_t$ denotes static and dynamic states of node $n$ at time $t$.
Unlike in machine translation, the state of nodes in the network status changes dynamically with shuttles movement; thus, we need to consider both static and dynamic states for each node.
Also, we let $y_t$ denote a node to be visited at time $t$ and $Y_t=\{y_1, \dots, y_t\}$ with $Y_0 = \emptyset$.
Then to select a next node to visit $y_{t+1}$, we are interested in learning $\phi(y_{t+1}|x_t, Y_t)$.

However, a set of nodes in the network does not convey any sequential information. Therefore, it is common in literature \cite{nazari2018reinforcement}, to omit recurrent neural network for encoding. Indeed, due to the sparse nature of networks, graph embedding is deployed in encoder to build their continuous vector representation as they suit better for statistical learning \cite{perozzi2014deepwalk}. The following equation describes embedding for each $n \in \Nc$:
\begin{align}
\bar{x}^n_s & = b^s + W^sx^n_s\\
\bar{x}^n_{d_t} & = b^d + W^dx^n_{d_t}
\end{align}
where, $\bar{x}^n_s$ and $\bar{x}^n_{d_t}$ are embedded static and dynamic states of node $n$ at time $t$ and $b, W$ represent the trainable parameters of a neural network. We denote by $\bar{x}^n_t = (\bar{x}^n_s;\bar{x}^n_{d_t})$ concatenation of embedded static and dynamic states of nodes.

For decoding we use recurrent neural networks (RNN), that takes static state of the last visited node and stores the sequence as follows:
\begin{equation}
h_t = W^hf(h_{t-1}) + W^{x}\bar{x}^n_s
\end{equation}
where $h_t$ is a memory state of RNN, $f$ represents nonlinear transformation and $x^n_s$ is a static state of node $n$ visited at time $t$. Trainable weight matrices $W^h$ and $W^{x}$ represent connections between hidden state to hidden state and hidden state to an input respectively. Note in our implementations, we use a LSTM cell as RNN.

In addition to encoder and decoder, we also utilize content based attention mechanism as in \cite{nazari2018reinforcement}. Content based attention tries to mimic associative memory and is designed to handle cases when an input to the sequence-to-sequence model is a set \cite{vinyals2015order}. In particular, the current state of an urban network is coupled with the memory state of RNNs about the sequence to calculate an alignment vector $c_t$ that assigns the probabilities of nodes to visit next:
\begin{align}
& u^n_t = v \tanh(W(\bar{x}^n_t;h_t)) && \forall n \in \Nc \\
& c_t = \text{softmax}(u_t)
\end{align}
where $v$ and $W$ are trainable weight matrices.

For the problem under study, we define the static state of nodes as their location coordinates and the initial charging levels of EVs at supplier nodes.
Even though the charging levels of EVs will change as EVs are taken to charging stations, only their initial values determine charging times.
Therefore, we consider them as a static state of nodes.
For a dynamic representation of the states of nodes, we use the number of EVs, the number of drivers in a shuttle, and the distance from the current node to other nodes.
Our experimental studies show that passing distance information as a dynamic state of nodes substantially reduces training time.
Figures \ref{fig:rl_overview} summarizes the sequence-to-sequence model of the shuttle routing problem used in the actor network.
In routing a fleet of shuttles, we also deploy a single actor network, where a sequence of visited nodes $Y_t$, includes nodes visited by all shuttles up to time $t$.

\subsection{Reward Function}
Reward function along with sets of available actions reflects our aim to maintain the feasibility and efficiency of routing decisions. Since the shuttle routing problem considers both charging and relocation of EVs, reward function must not only reflect traveling times between nodes, but also include waiting times. Therefore, we define reward function as the negative of total time spent in the system starting when a shuttle or a fleet of shuttles leaves a depot and ending when all shuttles are returned back to the depot with all drivers after fulfilling all demander nodes. Then our aim is to maximize the negative of total time spent in the system denoted by $R$. More formally we define reward function as follows, using immediate rewards $r_t$:
\begin{equation}
R = \sum_{t=1}^{T}r_t
\end{equation}
where \[ r_t =
\begin{cases*}
-\tau(n_{t-1}, n_{t}) & \text{if} $n_{t-1}\neq n_t$ \\
-w_t & \text{if} $n_{t-1} = n_t$  \\
\end{cases*} \]%
and $\tau_t$ is traveling time and $w_t$ is waiting time at time $t$.

\subsection{Training Algorithm}

In training, we are interested in finding policy parameters $\bar{\theta}$ that maximize the total expected reward:

\begin{equation}
\bar{\theta} = \argmax_{\theta} \mathbb{E}_{\pi_\theta} [ R ].
\end{equation}
Given the state of network $X$, we can write:
\begin{equation}
J(\theta|x) = \mathbb{E}_{\pi \sim p_{\theta}(\cdot|x)} [ R(\pi|x) ]
\end{equation}
and
\begin{align}
& \nabla_\theta J(\theta|x) = \mathbb{E}_{\pi}[A^\pi \nabla_\theta \log p_{\theta}(\pi|x) ] \\
& A^\pi = R(\pi|x)-V(x_0).
\end{align}
We use the REINFORCE algorithm with a baseline \cite{williams1992simple}, which is the value of the initial state of an urban network estimated by a critic with trainable parameters $\theta_c$.
Algorithm \ref{alg:multi} represents our training procedure, where the actor network with trainable parameters $\theta_a$ represents policy $\pi$. In a batch training setting, the batch of instances is generated by a data generator. Instances are passed through the simulator. Then, the actor network produces probabilities of nodes to be visited by shuttles at each time step, and the simulator is updated accordingly until the entire episode is finished. Then with the received total reward for the selected actions, the parameters of the actor and critic networks are updated. Unlike in the existing literature \cite{zulqar}, the algorithm does not require splitting an urban network into sub-clusters for each shuttle, but instead deploys all shuttles to serve the whole network. Also, utilizing a central controller that observes the entire urban network state along with the masking scheme in the simulator allows efficiently exploring joint action of all shuttles. For instance, if a node has been assigned to be visited by a shuttle, then that node is masked for other shuttles.

\begin{algorithm}
	\caption{Training Algorithm }
	\label{alg:multi}
	\begin{algorithmic}[1]
		\State {Initialize network parameters $\theta^{a}$ and $\theta^c $ for actor and critic networks respectively. Set the maximum number of epochs, a batch size and the maximum number of steps denoted as $M_\text{epochs}$, $M_\text{epis}$ and $T$ respectively;}
		\For{epochs = 1 to $M_\text{epochs}$ }
		\State{Reset gradients $d \theta^a, d \theta^c$;}
		\For{$m$ = 1 to $M_\text{epis}$}
		\State{data $\sim$ DataGenerator($\rho$);}
		\State{$x^m_0$, $\Ac_0$ = simulator.reset(data);}
		\State{Add $x^m_0$ to $X_0$, set $R^m=0$, set L to $\Ic$;}
		\For{t=0 to $T$}
		\ForEach {$i \in L $}
		\State $a^i_t$, $p^i_t$  = actor network($x_t$, $\Ac^i_t$);
		\State Store $p^i_t$ in $p^m$, remove $a^i_t$ from $\Ac_t$;
		\EndFor
		\State $x_{t+1}$, $\Ac_{t+1}$, $r_t$, $t_c$ = simulator.step($a_t$);
		\State Empty set $L$;
		\ForEach {$i \in \Ic $}
		\If{$a^i_t$ is complete at $t_c$}
		\State add $i$ to $L$
		\Else
		\State $a^i_{t+1}=a^i_t$ and remove $a^i_t$ from $\Ac_{t+1}$
		\EndIf
		\EndFor
		\State $R^m = R^m + r_t$;
		\EndFor
		\State{calculate $V^m(x^m_0; \theta_c)$ using critic}
		\EndFor
		\State{$d \theta_a =\frac{1}{M_\text{epis}}\sum_{m=1}^{M_\text{epis}} (R^m-V^m(x^m_0; \theta_c))\nabla_{\theta^a}\log p^m$;}
		\State{$d \theta_c = \frac{1}{M_\text{epis}}\sum_{m=1}^{M_\text{epis}} \nabla_{\theta_c}(R^m-V^m(x^m_0; \theta_c))^2$;}
		\EndFor
	\end{algorithmic}
\end{algorithm}

\section{Computational Studies}\label{sec:na}
\subsection{Data Generation and Configurations}
We consider a $1\times1$ square mile network consisting of demander, supplier, and charger nodes. We first specify the total number of nodes in the network and the number of demander and charger nodes. We sample x, y coordinate of each node from a uniform distribution with values ranging from 0 to 1. Similarly, we sample demander, charger, and supplier nodes from a uniform distribution. For each supplier node we set the initial charging levels of EVs randomly between 1 and 5. We assume that EVs do not need charging and can be directly taken to demander nodes if their charging levels exceed 3. Otherwise, EVs first need to be taken to charger nodes, where all of them are charged until the charging level of 5 is reached. For each charging level, we assign the charging time equal to the average traveling time between all pairs of nodes in the network. We do not consider discharging rates in the movements of EVs, while we assume the constant velocity for EVs equal to 45 miles/hour. 

Computational experiments are conducted with 2 Intel Xeon E5-2630 2.2 GHz 20-Core Processors, 32 GB RAM, and the Ubuntu 18.04.4 LTS operating system. 
All implementations are done in Python 3.7 using PyTorch 1.5. Our implementations of the critic network have similarities to the actor network structure except for the use of LSTM. We first embed the initial static state of the urban network using 1D convolution networks and then pass it to the attention mechanism. We pass the output of the attention mechanism through a sequence of feed-forward networks to obtain the final estimate for a value function.
Table \ref{tb:hp} represents the hyperparameters used for training, which are the same as in \cite{nazari2018reinforcement}.
We train RL agents on networks of various sizes and difficulty levels. For each problem class defined by the size of a network, we consider instances with three different levels of difficulty. Cases when there is an abundant presence of charging stations than the number of EVs requiring charging we call \emph{easy} instances. Similarly, cases when there is an exact number of charging stations as the number of demander nodes we call \emph{medium} difficulty instances. Finally, in cases when there is a less number of charging stations than the number of demander nodes, we call them \emph{hard} instances. The descriptions of difficulty levels are found in Table \ref{tb:instances}.
\begin{table}[]
	\caption{Hyperparamter values} 
	\centering
	\label{tb:hp}
	\begin{tabular}{@{}llll@{}}
		\toprule
		Conv1D, LSTM hidden dim &128 &
		Conv1D kernel size          & 1      \\
		Critic, linear hidden dim       & 128       &
		Learning rate actor, critic            & $10^{-4}$    \\
		 \bottomrule
	\end{tabular}
\end{table}

\begin{table*}[]
	\caption{Difficulty levels description, where $De$, $Ch$, $Su$, and $Su^{'}$ denote the set of demanders, chargers, suppliers, and suppliers with EVs that require charging, respectively.}
	\label{tb:instances}
	\centering
	\begin{tabular}{rrrrrrrrrrrrr}
		\toprule
		& \multicolumn{4}{c}{Easy}  & \multicolumn{4}{c}{Medium}   & \multicolumn{4}{c}{Hard}  \\
		\cmidrule(lr){2-5} \cmidrule(lr){6-9}  \cmidrule(lr){10-13}
		$|\Nc|$& $|De|$  & $|Ch|$ & $|Su|$  & $|Su^{'}|$ & $|De|$  & $|Ch|$ & $|Su|$  & $|Su^{'}|$ & $|De|$  & $|Ch|$ & $|Su|$  & $|Su^{'}|$ \\
		\midrule
		23  & 7  & 7  & 8  & 4          & 7  & 7  & 8  & 8          & 8  & 6  & 8  & 8          \\
		50  & 16 & 16 & 17 & 8          & 16 & 16 & 17 & 17         & 17 & 15 & 17 & 17         \\
		100 & 33 & 33 & 33 & 16         & 33 & 33 & 33 & 33         & 33 & 32 & 34 & 34     \\
		\bottomrule
	\end{tabular}
\end{table*}

\subsection{RL agents and Benchmarks}
We train three types of agents using the proposed RL models.
The first agent denoted as \emph{gen-RL} is trained on all three difficulty levels, but on a fixed network size.
The second agent denoted as \emph{net-RL} is trained on networks of various sizes, but it is tailored to a specific difficulty level.
The last agent denoted as \emph{RL} is trained on a fixed network size and on a specific difficulty level.
For our computational studies, we consider a benchmark from \cite{zulqar}. %
The benchmark model denotes as \emph{Sim} represents a joint model that solves the EVs relocation and the shuttle routing problems simultaneously. %
To solve multi-shuttle routing problems, the heuristic splits an urban network into some clusters and solves a single-shuttle routing problem for each cluster. However, there are some limitations to the method. One of them is related to the inflexibility of the solutions when drivers that have been dropped off from one shuttle cannot be picked up by other shuttles. Another disadvantage is related to charger nodes. The heuristic can only handle cases when the number of charger nodes is not less than the number of EVs that must be charged. %

\subsection{Results on Random Instances}

\begin{figure}
	\centering
	\includegraphics[scale=0.45]{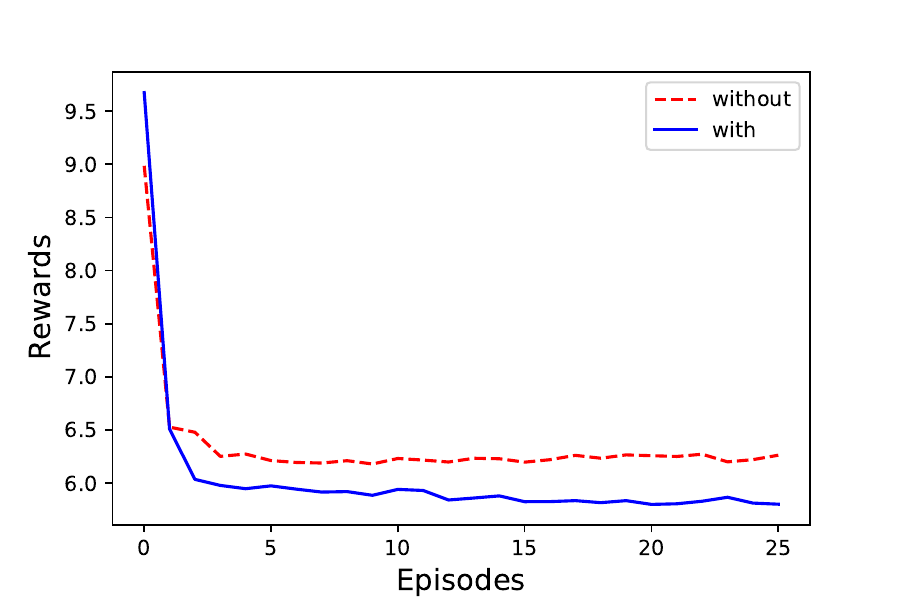}
	\caption{Training rewards with and without distance as an input}
	\label{fig:train}
\end{figure}

Figure \ref{fig:train} shows training rewards for the multi-shuttle problems on the network with 23 nodes and 3 drivers. Overall, training time depends on the network size, its structure, and the features passed to the actor network. Using distance information from the current node to other nodes in the actor network results in better rewards compared to when not passing such information.

To compare different RL agents' performances, we conduct experiments on various network sizes and the degree of difficulty of instances and measure the mean of the total time spent in the system out of 128 instances.
Table \ref{tb:rl_models} shows the experiments' results.
In most instances, an RL agent trained on a specific size and a specific instance difficulty level tends to perform the best.
We observe that net-RL agents, trained on various network sizes, tend to perform better on larger network sizes, while gen-RL agents, trained on various difficulty levels, can be competitive on medium-sized networks.
As the network size increases, the results show that using net-RL and gen-RL agents can be beneficial.
For the rest of the experiments, we use RL agents.

\begin{table*}[]
	\caption{Comparison of RL agents in terms of total time spent in the system, the average of 128 test instances are reported. In bold are the best results.}
	\label{tb:rl_models}
	\centering
	\begin{tabular}{@{}rrrrrrrrrrrr@{}}
		\toprule
		&   &          & \multicolumn{3}{c}{Easy} & \multicolumn{3}{c}{Medium} & \multicolumn{3}{c}{Hard} \\
		\cmidrule(lr){4-6} \cmidrule(lr){7-9}  \cmidrule(lr){10-12}
		$|\Nc|$ & $|\Ic|$ & $|Dr|$ & net-RL               & gen-RL            & RL               & net-RL               & gen-RL        & RL            & net-RL               & gen-RL          & RL             \\
		\midrule
		23       & 1     & 3  & 9.29     & 8.34  & \textbf{7.70}  & 14.63    & 11.75 & \textbf{10.27} & 16.01    & 13.73 & \textbf{12.32} \\
		& 2     & 3  & 6.01     & 5.79  & \textbf{5.40}  & 7.93     & 8.45  & \textbf{6.93}  & 8.89     & 9.00  & \textbf{8.34}  \\
		& 3     & 2  & 5.48     & 5.28  & \textbf{5.21}  & 7.02     & 7.58  & \textbf{6.38}  & 8.33     & 8.11  & \textbf{7.79}  \\
		50       & 1     & 3  & 14.97    & 13.96 & \textbf{13.77} & 20.35    & 19.36 & \textbf{17.93} & 22.60    & 20.05 & \textbf{18.92} \\
		& 2     & 3  & 8.54     & \textbf{8.21}  & 8.41  & 11.81    & \textbf{10.81} & 11.23 & 12.15    & 11.76 & \textbf{11.96} \\
		& 3     & 2  & 7.22     & 6.90  & \textbf{6.89}  & 9.58     & 9.41  & \textbf{9.23}  & 10.33    & 9.89  & \textbf{9.77}  \\
		100      & 1     & 3  & 23.16    & 22.98 & \textbf{22.18} & \textbf{30.62}    & 32.33 & 30.67 & \textbf{31.53}    & 32.30 & 36.67 \\
		& 2     & 3  & \textbf{12.91}    & 14.25 & 12.92 & 17.55    & 18.42 & \textbf{17.54} & \textbf{17.21}    & 18.79 & 17.90 \\
		& 3     & 2  & 10.23    & \textbf{10.21} & \textbf{10.21} & 13.39    & 13.73 & \textbf{13.33} & 13.69    & \textbf{13.67} & 14.94 \\ \bottomrule
	\end{tabular}
\end{table*}

\begin{table*}[]
	\caption{RL model vs. the heuristic optimization in terms of total time spent in the system, the percentages of winning instances and computational time in seconds, the average of 128 test instances are reported. In bold are the best results.}
	\label{tb:results_time}
	\centering
		\begin{tabular}{@{}rrrrrrrrrrrrrrrrr@{}}
			\toprule
			&   &    & \multicolumn{5}{c}{Easy}                                          & \multicolumn{5}{c}{Med}                                           & \multicolumn{4}{c}{Hard}                               \\ 
			
			\cmidrule(lr){4-8} \cmidrule(lr){9-13}  \cmidrule(lr){14-17}
			
			&   &    & \multicolumn{2}{c}{Mean} & Win \%  & \multicolumn{2}{c}{Time, s} & \multicolumn{2}{c}{Mean} & Win \%  & \multicolumn{2}{c}{Time, s} & \multicolumn{2}{c}{Mean} & \multicolumn{2}{c}{Time, s} \\
			
			\cmidrule(lr){4-5} \cmidrule(lr){6-6} \cmidrule(lr){7-8} \cmidrule(lr){9-10} \cmidrule(lr){11-11} \cmidrule(lr){12-13}
			\cmidrule(lr){14-15} \cmidrule(lr){16-17}
			
			$|\Nc|$ & $|\Ic|$ & $|Dr|$ & Sim         & RL         & RL-Sim   & Sim           & RL          & Sim         & RL         & RL-Sim   & Sim           & RL          & Sim        & RL          & Sim          & RL           \\
			\midrule
			23       & 1 & 3  & 8.81        & \textbf{7.70}       & 85.94\%  & 7.07          & 0.01        & 12.39       & \textbf{10.27}      & 94.53\%  & 13.98         & 0.02        &  --          & 12.32       & --             & 0.02         \\
			& 2 & 3  & 5.72        & \textbf{5.40}       & 65.63\%  & 1.61          & 0.04        & 7.43        & \textbf{6.93}       & 73.44\%  & 3.31          & 0.04        & --           & 8.34        & --             & 0.09         \\
			& 3 & 2  & 5.27        & \textbf{5.21}       & 51.56\%  & 0.91          & 0.06        & 6.39        & \textbf{6.38}       & 48.44\%  & 1.71          & 0.05        & --           & 7.79        &  --            & 0.11         \\
			50       & 1 & 3  & 17.34       & \textbf{13.77}      & 96.09\%  & 48.43         & 0.05        & 24.59       & \textbf{17.93}      & 100.00\% & 98.61         & 0.05        & --           & 18.92       & --             & 0.06         \\
			& 2 & 3  & 9.19        & \textbf{8.41}       & 74.22\%  & 11.62         & 0.21        & 12.25       & \textbf{11.23}      & 73.44\%  & 23.47         & 0.16        & --           & 11.96       & --             & 0.25         \\
			& 3 & 2  & 6.96        & \textbf{6.89}       & 53.13\%  & 5.35          & 0.20         & 9.25        & \textbf{9.23}       & 50.00\%  & 10.75         & 0.21        &--            & 9.77        &--              & 0.35         \\
			100      & 1 & 3  & 34.20       & \textbf{22.18}      & 100.00\% & 152.15        & 0.16        & 45.97       & \textbf{30.67}      & 100.00\% & 599.36        & 0.17        &--            & 36.67       & --             & 0.26         \\
			& 2 & 3  & 16.11       & \textbf{12.92}      & 100.00\% & 66.13         & 0.44        & 21.63       & \textbf{17.54}      & 96.09\%  & 118.68        & 0.44        & --           & 17.90       & --             & 0.55         \\
			& 3 & 2  & 11.71       & \textbf{10.21}      & 86.72\%  & 28.23         & 0.92        & 15.63       & \textbf{13.33}      & 92.97\%  & 53.86         & 1.03        & --           & 14.94       & --             & 1.02         \\ \bottomrule
		\end{tabular}
\end{table*}

Table \ref{tb:results_time} illustrates the performance of the RL solutions with those of the heuristic optimization method, labeled Sim.
The reinforcement learning approach can solve all instances of the problem, while the optimization method can handle only easy and medium cases. Moreover, for easy and medium cases measured in the mean of total time spent in the system, the RL solutions perform better than the heuristic optimization solutions. We also note that the derived RL solutions do not solve for optimal relocation of EVs and are only based on predefined rules, while the optimization heuristic solves for both the shuttle routing and EV relocation problems.

Table \ref{tb:results_time} shows the performance comparison of Sim and RL models in terms of percentages of winning instances. For instance, in an RL-Sim pair comparison, the value of cells under the column indicates the percentages of instances when the RL model performed at least equally to Sim model out of 128 test instances. As shown in Table \ref{tb:results_time} the RL model performs better than the heuristic method in at least 50\% of all instances, except one instance.  

To show the generation of the instantaneous solutions using RL models, we measured computation time.
Table \ref{tb:results_time} demonstrates the computation time it takes to derive a solution under Sim and RL models. We report an average time to solve an instance out of 128 instances in total. The difference in deriving solutions between Sim and RL models increases up to 585 times in the case of a single shuttle routing in a network with 100 nodes for Easy instances.

We also compare the effects of the number of drivers and difficulty levels on the trained models. In particular, we train models with a specific number of drivers on easy, medium, and hard instances on a fixed network size and check these models' performances against the models with varying a number of drivers and difficulty levels. For example, in Table \ref{tb:train_test_multi} rows indicate the problems' configurations in testing and columns indicate the problems' configurations in training datasets. The cells corresponding to a row and column show the percentages of instances when a trained model outperformed the model specifically trained for a test dataset. As we observe, models trained on specific difficulty levels tend to perform better on similar instances with a different number of drivers compared to on test models with the same number of drivers, but different difficulty levels.

\begin{table}
	\caption{The number of drivers vs. difficulty levels.}
	\label{tb:train_test_multi}
	\centering
	\resizebox{\linewidth}{!}{
	\newcommand\items{6}   %
	\arrayrulecolor{white} %
	\noindent
	\begin{tabular}{cc*{\items}{|E}|}
		\multicolumn{1}{c}{} &\multicolumn{1}{c}{} &\multicolumn{\items}{c}{Trained On} \\ \hhline{~*\items{|-}|}
		\multicolumn{1}{c}{} &
		\multicolumn{1}{c}{} &
		\multicolumn{1}{c}{\rot{E, $dr=2$}} &
		\multicolumn{1}{c}{\rot{M, $dr=2$}} &
		\multicolumn{1}{c}{\rot{H, $dr=2$}} &
		\multicolumn{1}{c}{\rot{E, $dr=3$}} &
		\multicolumn{1}{c}{\rot{M, $dr=3$}} &
		\multicolumn{1}{c}{\rot{H, $dr=3$}}
		\\ \hhline{~*\items{|-}|}
		\multirow{\items}{*}{\rotatebox{90}{Tested On}}
		&E, $dr=2$  & 0   & 14.8  & 16.4  &53.9 & 3.1 & 10.9\\ \hhline{~*\items{|-}|}
		&M, $dr=2$   & 6.3   & 0  & 21.9  & 0   & 8.6  & 14.8 \\ \hhline{~*\items{|-}|}
		&H, $dr=2$  & 10.9   & 3.9   & 0 & 0   & 6.2  & 40.6  \\ \hhline{~*\items{|-}|}
		&E, $dr=3$  & 23.4   & 11.7  & 12.5 & 0   & 6.3  & 7.0  \\ \hhline{~*\items{|-}|}
		&M, $dr=3$   & 14.8   & 39.8  & 14.1 & 0   & 0  & 14.8  \\ \hhline{~*\items{|-}|}
		&H, $dr=3$  & 17.2   & 0.8   & 37.5  & 0   & 3.1  & 0 \\ \hhline{~*\items{|-}|}
	\end{tabular}}
	\arrayrulecolor{black} %
\end{table}

\begin{table}
	\caption{The Amsterdam dataset structure and RL model vs. Sim on the dataset in terms of total time spent in the system.}
	\label{tb:amsterdam}
	\centering
	\begin{tabular}{@{}rrrrrrr@{}}
		\toprule
		&   &    & \multicolumn{2}{c}{Weekdays} & \multicolumn{2}{c}{Weekends} \\ 
		\cmidrule(lr){4-5} \cmidrule(lr){6-7}
		$|\Nc|$ & $|\Ic|$ & $|Dr|$ & Sim           & RL           & Sim           & RL           \\
		\midrule
		170 & 1 & 5  & 420.08        & \textbf{416.23}       & \textbf{411.59}      & 433.63       \\
		    & 2 & 3  & 207.63        & \textbf{182.39}       & 224.88        		& \textbf{185.06}       \\
		    & 3 & 2  & 142.90        & \textbf{136.52}       & 153.38        		& \textbf{142.20}       \\ \bottomrule
	\end{tabular}
\end{table}

\begin{figure}
\begin{minipage}{\linewidth}
	\centering
	\includegraphics[width=0.85\textwidth]{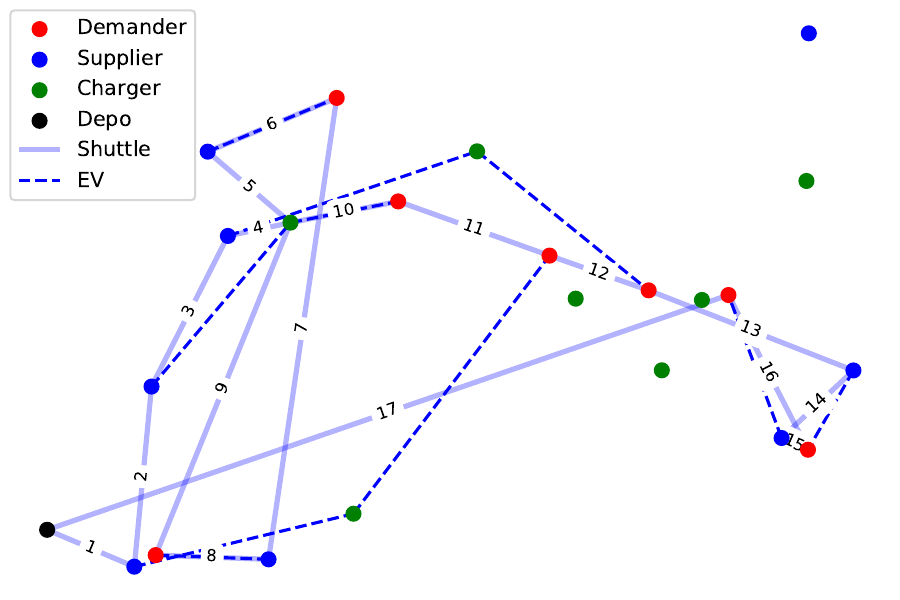}
	\captionof{figure}{Example solution for a single-shuttle case, $|\Nc|=23$, $|Dr|=3$ and $|\Ic|=1$.}
	\label{fig:policy}
\end{minipage}
\end{figure}

\begin{figure}
	\begin{minipage}{\linewidth}
		\centering
		\includegraphics[width=0.85\textwidth]{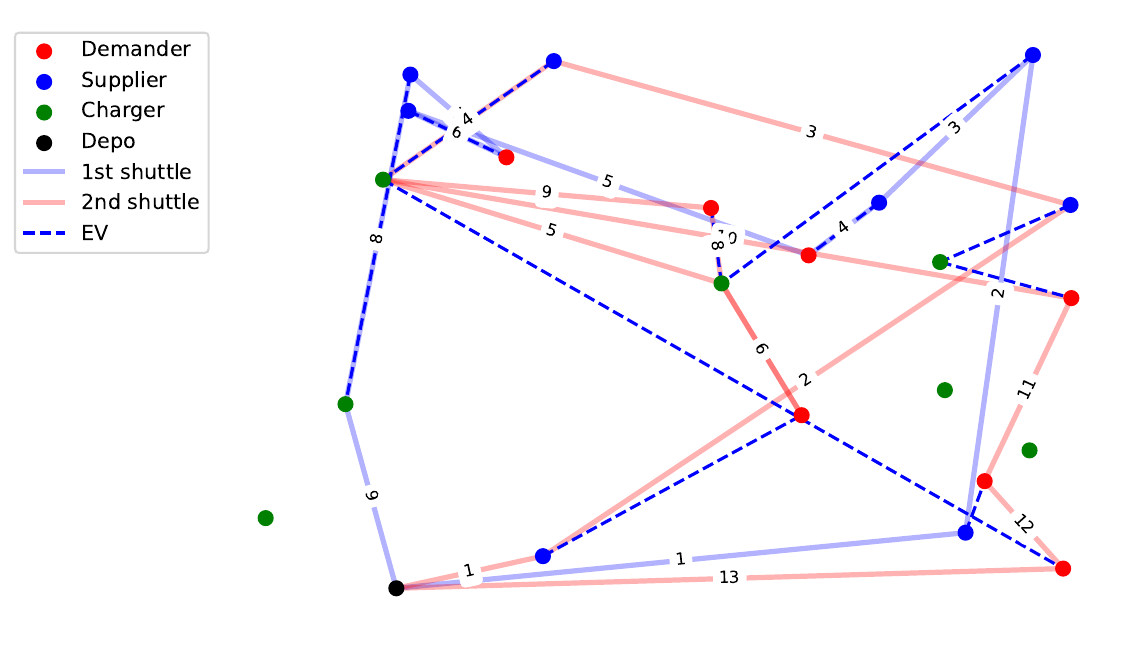}
		\captionof{figure}{Example solution for a multiple-shuttle cases, $|\Nc|=23$, $|Dr|=3$ and $|\Ic|=2$.}
		\label{fig:policy_multi}
	\end{minipage}
\end{figure}

The sample solution for a single-shuttle case, where 4 EVs in an urban network require charging, is shown in Figure \ref{fig:policy}. A shuttle with 3 drivers leaves the depot and visits supplier nodes first, followed by a charger node. By interchangeably visiting nodes thorough the network, the shuttle returns to the depot after picking up drivers from demander nodes. We can observe the versatility of the produced solutions by looking at the charging stations. For instance, a driver dropped off at the first visited supplier node relocates the EV to a charging station, waits there until the EV is charged, and then relocates it to a demander node. Only then the driver is picked up by a shuttle. In another example, the driver dropped off at the second visited supplier node is picked up immediately at a designated charging station by a shuttle.
Similarly, Figure \ref{fig:policy_multi} represents the sample solution for the case with 2 shuttles. Each shuttle visits supplier nodes first until it runs out of drivers. Then each of them interchangeably visits charger, supplier, and demander nodes and returns to the depot. The flexibility of the produced solutions can be observed when a driver originally dropped at the second visited supplier node by the first shuttle is picked up at a charging station by the second shuttle.

\subsection{Results on the Amsterdam Cases}

We also use real data of FFEVSS representing car2go operations in Amsterdam, the Netherlands, which was collected between May 5th and October 29th, 2016. From the actual data, we collect locations of supplier, demander and charger nodes and reduce the network by removing EVs that do not need relocation/charging. We also group the data into weekdays and weekends, which results in 14 and 12 instances for the respective groups that we use as test data for the experiments. To train an RL agent, we generate on the fly training data by sampling locations of nodes using weekdays data by extracting the CDF of the distribution for nodes' coordinates. We also assign node types randomly by following the similar structure observed in weekdays data. In particular, the training and testing data have 170 nodes with 71 supplier, 71 demander, and 27 charger nodes. Table \ref{tb:amsterdam} presents the results of the RL agent performance on the Amsterdam dataset. In all instances with except one, the RL agent performs better as compared to the heuristic; overall, RL performs 6.17\% better. The poor performance of the RL agent in the instance with a single shuttle and 5 drivers on weekends may be improved by training with more weekend data.

\section{Conclusion}\label{sec:conclusion}

This study solves the shuttle routing problem for FFEVSS. 
We consider a static network, in which a group of drivers is deployed to relocate EVs from supplier nodes to charger and demander nodes. 
We propose a reinforcement learning approach to learn routing policies for single-shuttle and multi-shuttle cases. The proposed solution methods allow solving the new class of problem instances while demonstrating improved results on instances solvable by existing methods in the literature. We also present several RL agents that generalize on various network structures or network sizes, and we demonstrate that the RL agent specifically trained on a network produces superior results. As future work, it is interesting to consider a dynamic network with the presence of customer demand to rebalance FFEVSS in the daytime.

\bibliographystyle{ieeetr}
\bibliography{reference}

\clearpage

\end{document}